# SoftReMish: A Novel Activation Function for Enhanced Convolutional Neural Networks for Visual Recognition Performance

Mustafa Bayram GÜCEN

Yildiz Technical University, Faculty of Arts and Sciences, Mathematics, 34220, İstanbul, Turkey

*Corresponding author e-mail: mgucen@yildiz.edu.tr

*Abstract*

In this study, SoftReMish, a new activation function designed to improve the performance of convolutional neural networks (CNNs) in image classification tasks, is proposed. Using the MNIST dataset, a standard CNN architecture consisting of two convolutional layers, max pooling, and fully connected layers was implemented. SoftReMish was evaluated against popular activation functions including ReLU, Tanh, and Mish by replacing the activation function in all trainable layers. The model performance was assessed in terms of minimum training loss and maximum validation accuracy. Results showed that SoftReMish achieved a minimum loss ($3.14 \times 10^{-8}$) and a validation accuracy (99.41%), outperforming all other functions tested. These findings demonstrate that SoftReMish offers better convergence behavior and generalization capability, making it a promising candidate for visual recognition tasks.

***Keywords:*** *Convolutional Neural Networks, Visual Recognition, Activation Functions*

## INTRODUCTION

Activation functions play a vital role in the performance of deep neural networks, especially by introducing nonlinearity and enabling complex feature learning(Dubey et al., 2022, Wang et al., 2020). Traditional activation functions such as ReLU and Tanh have gained popularity due to their simplicity and computational efficiency. Despite their widespread use, these functions exhibit certain limitations ReLU is prone to the "dying neuron" problem, while Tanh can suffer from saturation, leading to vanishing gradients and slower convergence. Therefore, researchers have proposed alternative activation functions to mitigate these issues and enhance model performance (Zhou et al., 2020). While traditional functions such as ReLU, tanh and more recently Mish activation function have shown remarkable success, each comes with its own drawbacks in terms of smoothness, saturation and gradient behavior(Misra, 2019).

In this context, the present study introduces SoftReMish, a novel activation function that integrates the advantages of both smooth and non-linear transformations. The experimental results demonstrate that SoftReMish can effectively address the shortcomings of traditional functions and offers a more robust solution for complex learning tasks. Inspired by these observations, we present a new activation function called SoftReMish. This function combines the smoothness of Mish activation function with advanced nonlinear scaling that allows for better gradient propagation in deep networks (Misra, 2019).

We conducted extensive experiments on the MNIST dataset to evaluate the effectiveness of SoftReMish (LeCun et al., 1998). Our results show that networks using SoftRemish outperform those using ReLU and even Mish in terms of classification accuracy and training stability. The ability of the proposed function to maintain the gradient flow while preserving nonlinearity makes it a promising candidate for further investigation in more complex vision and language tasks.

The initial findings presented in this study suggest that a new perspective in activation function design is feasible. With further evaluation on different datasets and network architectures, the potential impact of SoftReMish in deep learning can be more thoroughly assessed. In this context, the proposed function is expected to be beneficial across a wide range of applications.

## MATERIAL AND METHODS

### Material

The material used in this study is the MNIST dataset, which consists of 60,000 training and 10,000 test images of handwritten digits ranging from 0 to 9. Each image is 28×28 pixels in size and in grayscale format. The dataset was selected due to its standardized structure, wide usage in benchmarking neural network performance, and suitability for evaluating the effectiveness of activation functions in classification tasks (LeCun et al., 1998).

### Methods

*Activation Functions and SoftReMish*

Activation functions are essential components of deep neural networks, as they introduce non-linear transformations that allow the model to capture and represent complex patterns within the data. Among these, the Rectified Linear Unit (ReLU) is widely adopted in deep learning architectures because of its computational efficiency and straightforward implementation, making it effective for training deep models (Nair and Hinton, 2010). It is mathematically defined as:

$$f(x) = max(0, x)$$

This function returns the input value directly if it is positive; otherwise, it returns zero. ReLU helps mitigate the vanishing gradient problem encountered in deep networks and accelerates convergence during training(Krizhevsky et al., 2012). However, one of its known drawbacks is the "dying ReLU" problem, where neurons can become inactive if the input remains negative over time.

The Tanh activation function is a conventional, smooth and differentiable function that maps input values to the range (-1, 1). It is defined as:

$$f(x) = tanh(x) = (e^x - e^{-x}) / (e^x + e^{-x})$$

Compared to the sigmoid function, Tanh is zero-centered, which often leads to faster convergence. It is particularly useful in problems requiring normalized outputs. However, like the sigmoid, Tanh suffers from the vanishing gradient problem for large input values, as the gradients approach zero at both ends of the function.

The Mish activation function is a non-monotonic, smooth, and self-regularizing function that has demonstrated improved performance in various deep learning tasks. It is defined as:

$$f(x) = x \cdot tanh(ln(1 + e^x)) = x \cdot tanh(softplus(x))$$

where the function $tanh(x) = (e^x - e^{-x}) / (e^x + e^{-x})$ (Misra, 2019). The function softplus is convex defined $softplus(x) = ln(1 + e^x)$ (Dugas et al., 2009). Mish retains small negative values and avoids hard zero cut offs, unlike ReLU. Its smooth nature allows better gradient flow during backpropagation, potentially leading to improved model generalization and training dynamics.

Common activation functions such as Tanh, ReLU, and more recently Mish have demonstrated different strengths and limitations depending on the task and architecture. ReLU is computationally efficient and effective in many cases but suffers from the "dying ReLU" problem, where neurons can become inactive. Tanh offers smoother gradients but can lead to vanishing gradient issues in deep networks. Mish, a self-regularized non-monotonic activation function, has shown promising results in preserving gradient flow and improving generalization.

SoftReMish is a novel activation function proposed as a smooth and bounded alternative that combines the benefits of Mish and exponential-based transformations. Mathematically defined as

$$f(x) = x.tanh(ln(1 + exp(\alpha x)))$$

SoftReMish introduces a steeper activation response in high-value domains while maintaining stability around the origin. In this study, alpha is set to 2 in order to perform the numerical calculations. This property helps in enhancing the learning capacity of the network, especially in early training phases. In our experiments, we compare the performance of SoftReMish against traditional and modern activation functions using the MNIST dataset, evaluating its effect on training convergence and accuracy.

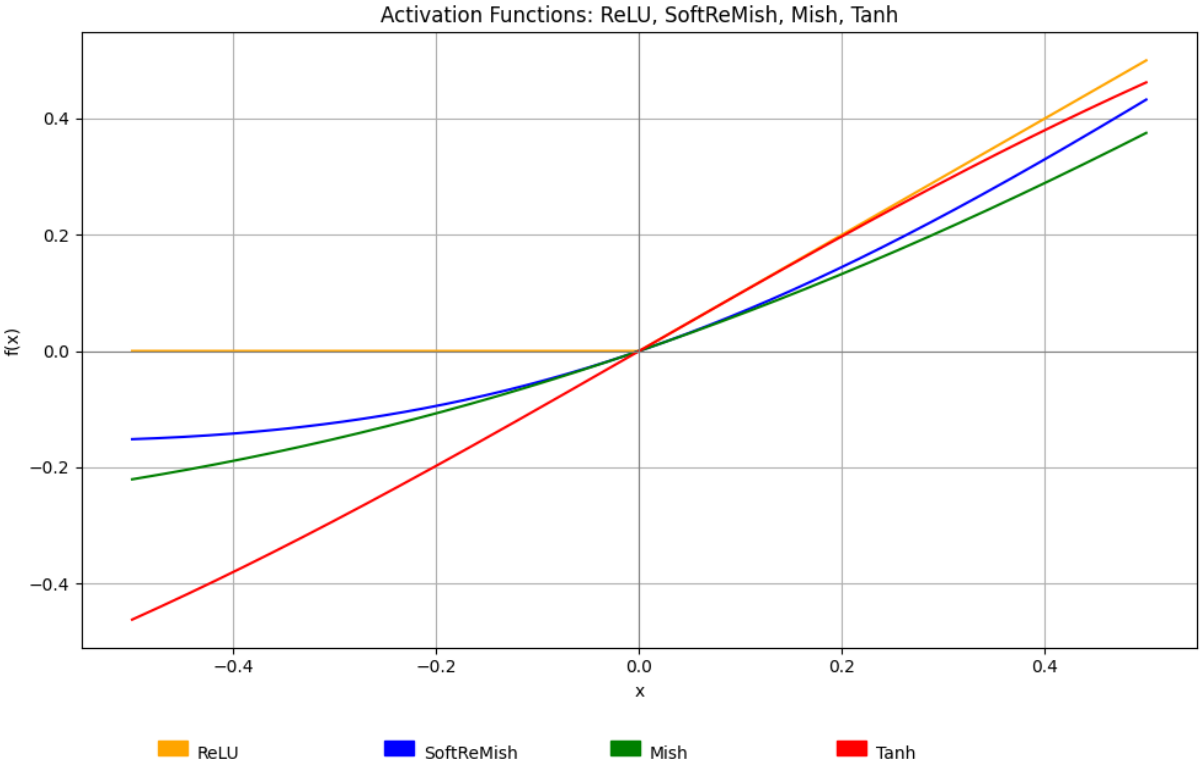

**Figure 1.** Graphs of Activation Functions ReLU, Tanh, Mish and SoftReMish

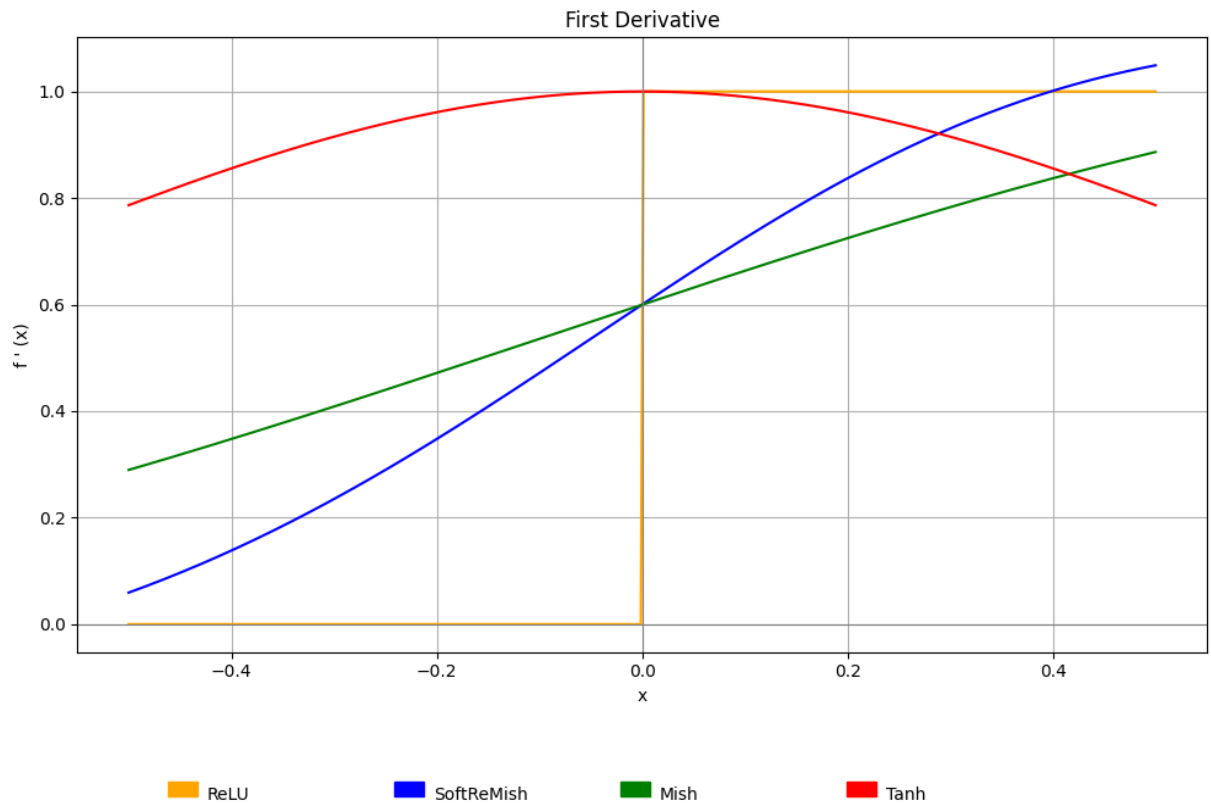

**Figure 2.** First Derivatives of Activation Functions ReLU, Tanh, Mish and SoftReMish

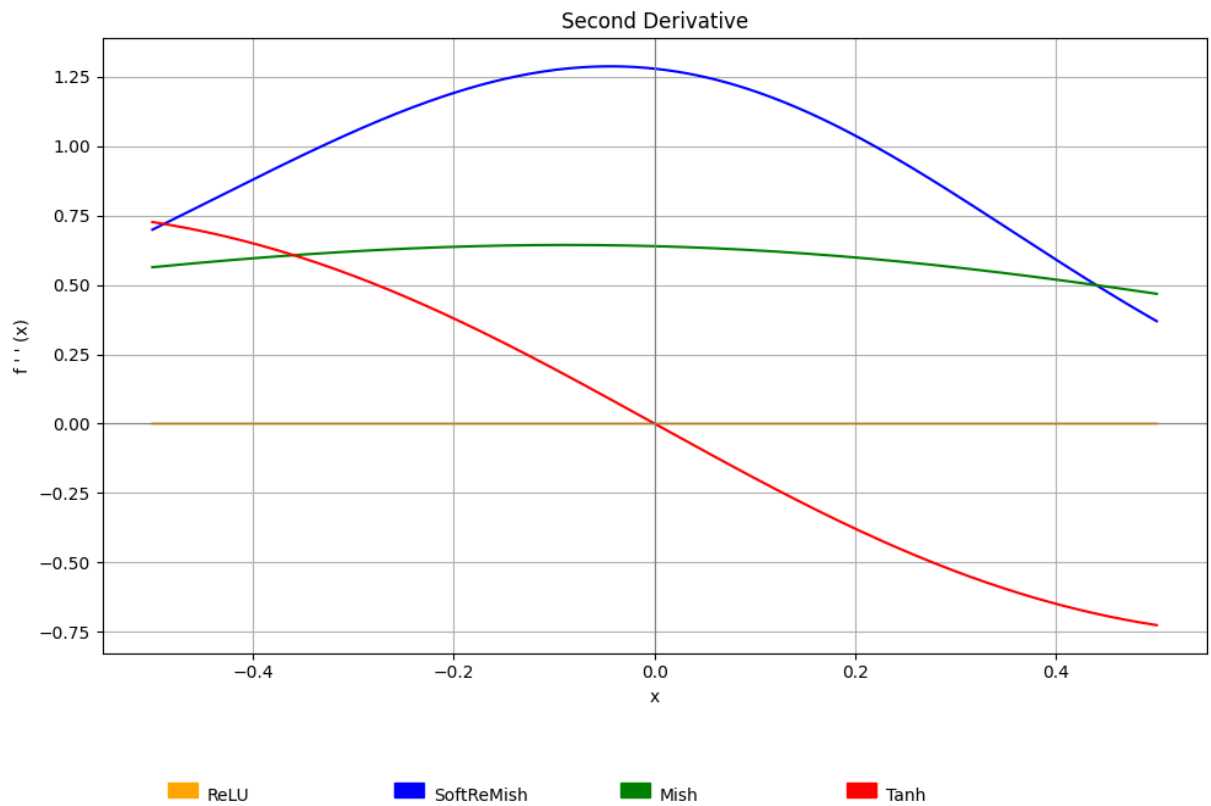

**Figure 3.** Second Derivatives of Activation Functions ReLU, Tanh, Mish and SoftReMish

*Convolutional Neural Networks*

Convolutional Neural Networks are the class of deep neural network models especially well-adapted for image classification tasks(Wang et al., 2020). They leverage convolutional layers to automatically learn spatial hierarchies of features, reducing the need for manual feature engineering. A typical CNN architecture consists of convolutional layers followed by pooling layers which downsample the spatial dimensions and dense layers for classification. By sharing weights across space and using local receptive fields, CNNs achieve parameter efficiency and translation invariance. Model explanation and layers defined detailed.

Conv2D (Convolutional Layer)

Convolutional layers are fundamental components of CNNs, especially effective in image processing tasks. This layer extracts local features from the input image. In the first convolutional layer, 32 filters of size 3×3 are applied to perform the convolution operation. The filters slide across the image in small regions, learning features like edges, textures, or patterns. This process extracts low-level features from the image.

The activation function is used to introduce non-linearity, which allows the model to learn complex relationships. Specifically, it outputs zero for any negative input, enabling the network to learn non-linear features.

This layer takes an input image of size 28×28×1 and begins learning higher-level features by performing convolutions with each filter.

MaxPooling2D (Max Pooling Layer)

The MaxPooling2D layer is commonly utilized following convolutional layers to decrease computational load while maintaining critical features of the input. By applying a 2×2 pooling window, this layer extracts the highest value from each subregion, effectively downsampling the spatial dimensions of the feature maps. This dimensionality reduction not only lessens the number of parameters in the network thereby improving computational efficiency but also enhances the model's ability to handle minor spatial shifts in the input data, contributing to greater robustness.

Conv2D (Second Convolutional Layer)

The second convolutional layer processes the more abstract features extracted by the first convolutional layer. In this layer, 64 filters of size 3×3 are used. The goal of this layer is to learn more complex and higher-level features from the output of the previous layer.

This layer enables the neural network to learn deeper, more abstract representations of the inputs with the increased number of filters allowing for more nuanced feature extraction.

MaxPooling2D (Second Max Pooling Layer)

The second max pooling layer performs a similar operation to the first, reducing the spatial dimensions of the output further and helping to minimize overfitting. This additional pooling step ensures that the learned features become increasingly abstract, with less emphasis on fine-grained spatial details and more focus on the structure of the input data.

Flatten Layer

The flatten layer is used to convert the 2D feature maps obtained from the convolutional and pooling layers into a 1D vector. After the convolutional and pooling layers, the output is still in a multi-dimensional form. The flatten layer takes this output and transforms it into a 1D vector so that it can be fed into fully connected layers for further processing.

This step is crucial because fully connected layers require a 1D vector as input, and this transformation allows the network to transition from learning spatial features to making classification decisions.

Dense (Fully Connected Layer)

The fully connected layer (dense layer) connects each neuron to every neuron in the previous layer. This layer allows the network to learn more abstract relationships from the features learned in earlier layers. In this model, the first dense layer contains 128 neurons and uses the ReLU activation function.

The dense layer enables the network to combine and manipulate learned features to make higher-level decisions. It plays an essential role in abstracting the learned information into more complex decision-making patterns.

Dense (Output Layer)

The final dense layer represents the output layer of the network. It contains as many neurons as the number of classes in the classification problem, which is 10 in this case (for a 10-class classification task, such as digit recognition in the MNIST dataset). The softmax activation function have been used in this layer.

The Softmax activation function transforms the model's raw output scores into a normalized probability distribution, where each output neuron corresponds to the likelihood of the input belonging to a specific class. This function guarantees that all output values fall within the range [0, 1], and that their sum equals 1, enabling probabilistic interpretation for classification tasks.

The described model follows a conventional Convolutional Neural Network (CNN) framework, which is widely employed in image classification applications. Convolutional layers serve to extract fundamental features from the input data, while pooling layers reduce spatial resolution and emphasize the most salient patterns. Subsequently, fully connected layers integrate these extracted features to perform the final classification. This architectural design has demonstrated strong performance in tasks such as handwritten digit recognition, facial recognition, and other visual classification challenges.

In this study, we implemented a CNN architecture with two convolutional and max pooling layers, followed by a fully connected layer and a final softmax output layer. We kept the overall architecture constant while varying the activation functions in each experiment to isolate their impact on performance. The models were trained and evaluated on the MNIST dataset, a standard benchmark for handwritten digit recognition. The results demonstrate how the choice of activation function can significantly affect validation loss, validation accuracy.

**RESULTS**

The comparative analysis of various activation functions demonstrated that the SoftReMish function outperformed the others in terms of both validation accuracy and validation loss. As shown in Table 1, the highest validation accuracy was achieved using the SoftReMish activation function with a value of 0.9941, followed by ReLU (0.9923), Tanh (0.9918), and Mish (0.9907). These results indicate that SoftReMish provides a slight yet significant improvement in classification performance compared to more conventional activation functions.

**Table 1.** Validation Accuracy Values Table

| Activation Function | Validation Accuracy |
| --- | --- |
| SoftReMish | 0.9941 |
| ReLU | 0.9923 |
| Tanh | 0.9918 |
| Mish | 0.9907 |

In addition to accuracy, validation loss values presented in Table 2 further support the superiority of the SoftReMish function. It yielded the lowest validation loss of $3.137582 \times 10^{-8}$, which is substantially lower than those observed for ReLU ($3.215818 \times 10^{-4}$), Tanh ($1.566297 \times 10^{-4}$), and Mish ($7.632310 \times 10^{-4}$). This remarkable reduction in validation loss signifies better generalization and model stability during training.

**Table 2.** Validation Loss Values Table

| Activation Function | Validation Loss |
|---|---|
| SoftReMish | $3.137582 \times 10^{-8}$ |
| ReLU | $3.215818 \times 10^{-4}$ |
| Tanh | $1.566297 \times 10^{-4}$ |
| Mish | $7.632310 \times 10^{-4}$ |

Overall, the findings of this study suggest that the proposed SoftReMish activation function can be a promising alternative to traditional functions such as ReLU, Tanh, and Mish, particularly in applications where high accuracy and low loss are critical.

Figure 4. illustrates the validation loss values corresponding to each activation function evaluated in the study. The SoftReMish activation function achieved the lowest validation loss with a value of $3.137582 \times 10^{-8}$, indicating a highly stable and well-generalized model. In comparison, traditional activation functions such as ReLU, Tanh, and Mish resulted in significantly higher loss values, with Mish yielding the highest at $7.632310 \times 10^{-4}$. The clear difference in loss values emphasizes the effectiveness of the SoftReMish function in reducing overfitting and enhancing model performance.

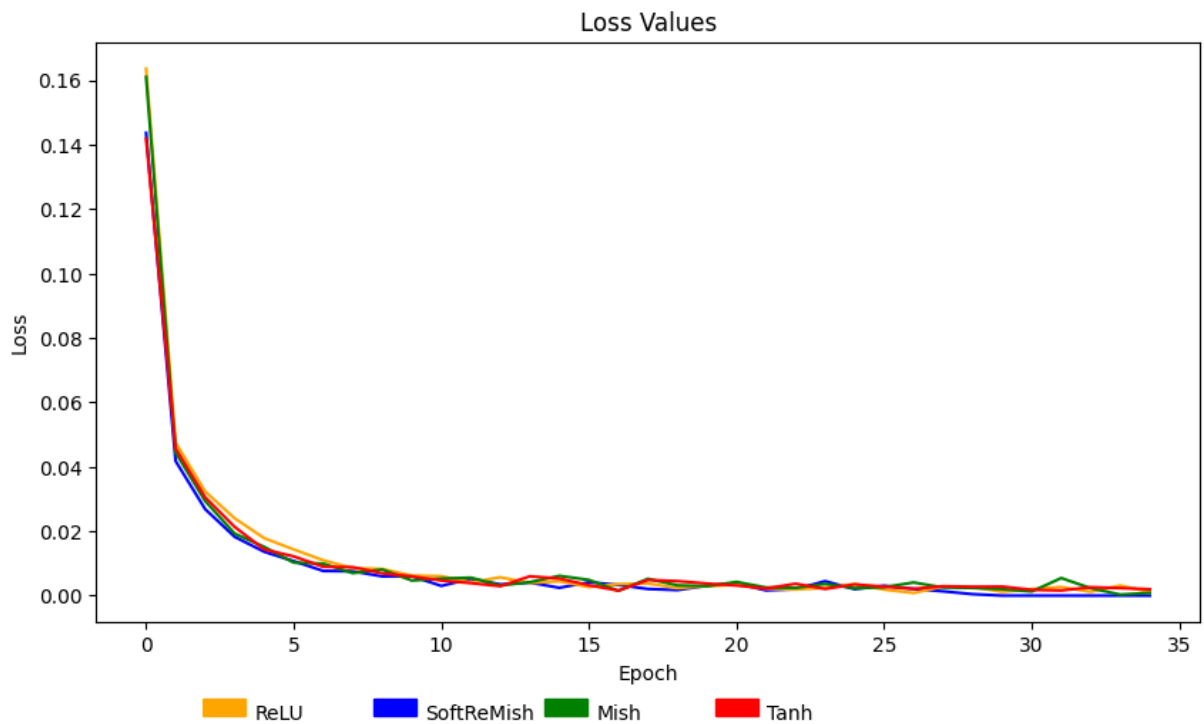

**Figure 4.** Validation loss values

Figure 5. presents the validation accuracy results for the tested activation functions. Among them, the SoftReMish activation function achieved the highest accuracy of 0.9941, surpassing ReLU (0.9923), Tanh (0.9918), and Mish (0.9907). These results confirm that SoftReMish not only minimizes loss but also contributes to a more accurate model. The improvements, although numerically small, are statistically and practically meaningful in high-precision tasks, reinforcing the advantage of using SoftReMish in deep learning models.

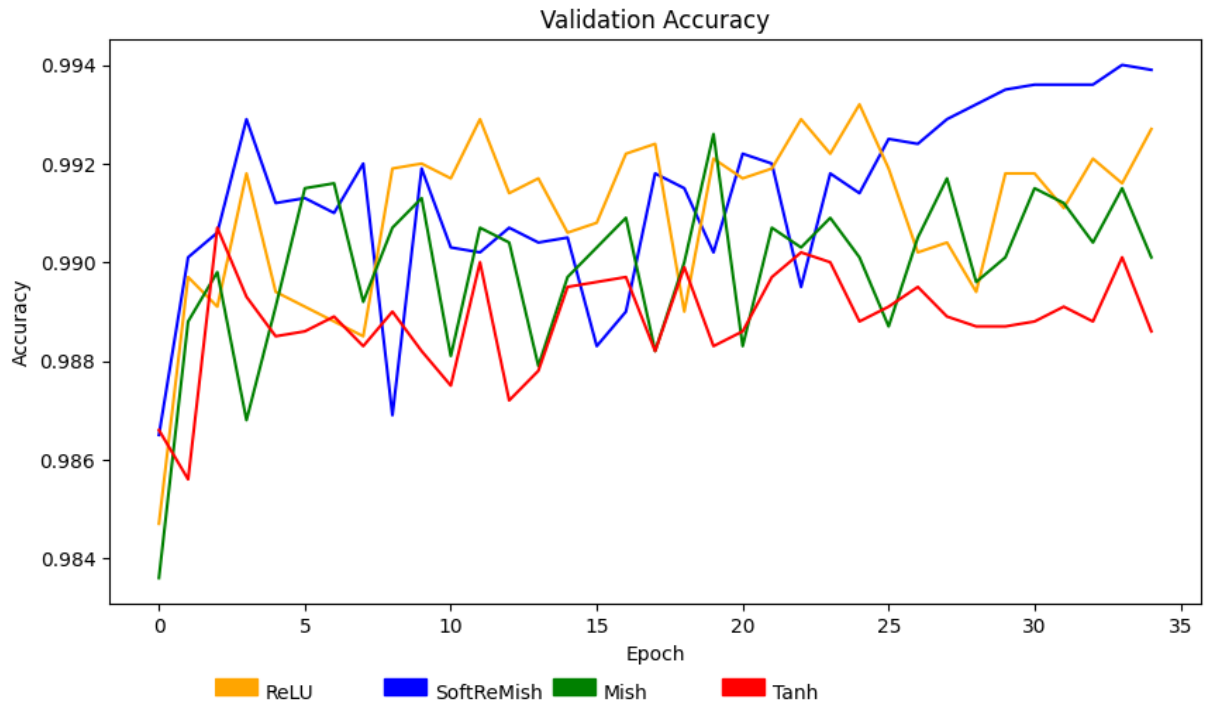

**Figure 5.** Validation accuracy values

All experiments were conducted using Python 3.11.12 on a Linux (kernel 6.1.123+) operating system. Model development and training were performed using an NVIDIA A100-SXM4-40GB GPU with CUDA 12.4 and NVIDIA driver version 550.54.15. The system was equipped with 83.48 GB of RAM and an x86_64 architecture CPU. GPU acceleration was utilized throughout training to optimize computation time. The libraries such as TensorFlow or PyTorch (depending on implementation) were used to construct and evaluate the CNN architecture for efficient experimentation. These results demonstrate that the SoftReMish activation function outperforms traditional activation functions in terms of both accuracy and loss. Therefore, SoftReMish can be considered a strong alternative for achieving higher accuracy and different generalization in deep neural networks.

**CONCLUSION**

In this study, the performance of four different activation functions—SoftReMish, ReLU, Tanh, and Mish—was evaluated based on their validation accuracy and validation loss in a deep learning model. The experimental results clearly indicate that the proposed SoftReMish activation function achieves superior results compared to the other functions tested. It provided the highest validation accuracy (0.9941) and the lowest validation loss ($3.137582 \times 10^{-8}$) suggesting that it leads to better model generalization and training stability.

Traditional activation functions like ReLU and Tanh have been widely used due to their simplicity and computational efficiency. However, their limitations, such as dying neurons in ReLU or saturation issues in Tanh, can hinder model performance. Therefore, alternative activation functions have been proposed to address these shortcomings (Zhou et al., 2020). The results of this study show that SoftReMish, by combining the strengths of both smooth and non-linear transformations, can overcome some of these limitations and offer a more robust alternative for complex learning tasks.

Moreover, the remarkably low validation loss achieved by SoftReMish indicates that it can reduce overfitting, an important concern in deep neural networks. This makes it particularly suitable for applications where both accuracy and generalization are critical, such as medical diagnostics, autonomous systems, and financial forecasting.

In conclusion, the SoftReMish activation function holds significant promise for improving deep learning model performance. Future studies could explore its applicability across different architectures and datasets to further validate its effectiveness and to optimize its design for various real-world tasks.

**APPENDİX**

**Conference Publication**